\documentclass[default,iicol]{sn-jnl}% Default with double column layout

\jyear{2021}%

\usepackage[T1]{fontenc}
\usepackage{lmodern}
\usepackage{microtype}
\usepackage{graphicx}
\usepackage[algo2e,ruled,vlined]{algorithm2e} 
\usepackage[caption = true]{subfig}
\usepackage{booktabs} % for professional tables
\usepackage{amsmath, amsfonts, color,setspace}
\usepackage{enumitem}
\setlist[itemize]{noitemsep, topsep=0pt}
\setlist[enumerate]{noitemsep, topsep=0pt}

\newcommand{\E}{\mathbb{E}}
\newcommand{\fhat}{\hat{f}}
\newcommand{\chat}{\hat{c}}
\begin{document}

\title{Using Image Transformations to Learn Network Structure} 
%%[Using Image Transformations to Learn Network Structure]



\author*[1]{\fnm{Brayan} \sur{Ortiz}}\email{brayanvo@amazon.com}

\author[1]{\fnm{Amitabh} \sur{Sinha}}\email{amitabsi@amazon.com}
\equalcont{These authors contributed equally to this work.}

\affil*[1]{\orgdiv{Modeling and Optimization}, \orgname{Amazon.com}, \city{Seattle}, \state{WA}, \country{USA}}

\abstract{Many learning tasks require observing a sequence of images and making a decision. In a transportation problem of designing and planning for shipping boxes between nodes, we show how to treat the network of nodes and the flows between them as images. These images have useful structural information that can be statistically summarized. Using image compression techniques, we reduce an image down to a set of numbers that contain interpretable geographic information that we call \textit{geographic signatures}. Using geographic signatures, we learn network structure that can be utilized to recommend future network connectivity. We develop a Bayesian reinforcement algorithm that takes advantage of statistically summarized network information as priors and user-decisions to reinforce an agent's probabilistic decision.}


\maketitle

\section{Introduction}\label{introduction}

\begin{figure}[h]
	\centering
		\vspace{-0.2in}
	\includegraphics[height=1in,width=1.5in]{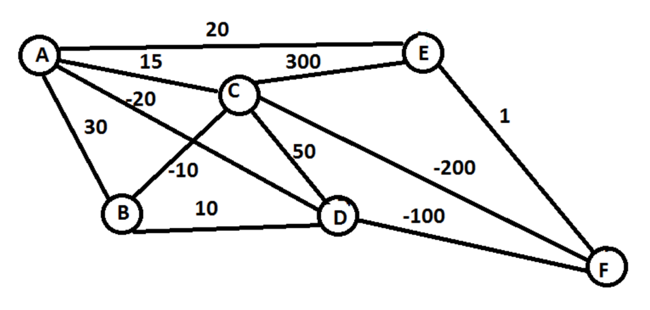}
	\caption{Mock topology and cost structure. }
	\label{fig:tsp}
		\vspace{-0.2in}
\end{figure}

In transportation and logistics companies, decisions have to be made frequently by operators on which warehouses or nodes need to be connected, i.e. the travelling salesperson problem (TSP). This problem has constraints and structure. We can abstract this down to a simpler scenario as in~Figure~\ref{fig:tsp}. We want to ship a package from $A$ to $F$ with minimal cost, where cost is shown on the edges. Nodes $C$, $D$, and $E$ are relatively well-connected to $F$, i.e. we are “positively rewarded” for taking $C\rightarrow F$ and $D\rightarrow F$ (cost is negative as in savings), while $E\rightarrow F$ is only \$1. Cost can be the shipping cost of each box or the cost of ordering trucks. Unfortunately, in practice, such a topology and structure of costs becomes obsolete either because of stochasticity or evolution: new nodes are added (new node $G$ either within the graph or outside of it), new connections ($A\rightarrow F$ or $B\rightarrow E$ directly, $A\rightarrow G$, etc.), and most importantly the cost structure is changing based on stochastic demand and pricing. 

This paper focuses on the decisions made by an operator choosing short-term network activity. In a short enough window, changes within a network are likely small, but identifying consistent connections and recommending new connections based on network activity is valuable. Airlines, for example, change connection structure only every so often. Our goal is to define a mathematical representation of a transportation network that allows experimentation of new nodes or connections by an operator that is making short-term decisions on which connections should be opened. Hence, the operator needs a set of changes that is feasible to process (small enough) and presented in order of importance (ranked).

Ideally, to choose connections that will be active (have boxes shipping), we can forecast the number of boxes flowing within all connections for a given network while respecting network or geographic constraints, such as node capacities or transit times. The traditional approach to network design and planning in operations is to consider global optimization techniques that solve for connectivity and flows while constrained for asset capacities \citep{kallrath2005solving, floudas2009review}. This global approach is useful for strategic or long-term decisions. Batch or campaign style planning is well-studied in industrial processes, but in transportation it could only be effective provided standing decision mechanisms are aligned, such as simultaneous launch of buildings, etc. \citep{kallrath1999concept, grunow2002campaign, berning2002integrated, mendez2002milp}. Overall, when we approach this problem by embedding unique node and time specific information, then the mathematical model does not generalize beyond that network state. If we can summarize network activity at some aggregate level while capturing network dependencies, then it would be possible to estimate the cost of new networks with added nodes and connections. 

For the TSP problem in general as represented by structural graphs, there are computationally expensive methods that leverage learned abstractions of historical graphs via deep neural networks to explore unseen instances of graph activity \citep{bello2016neural, khalil2017learning}. However, it would be difficult to experiment with such a method if the user has geographical properties in mind, such as location of a connection relative to a highly active node. In this paper, we show that these flows can be represented as images unique to each node and time point. Using compression techniques to reduce the dimension of the images, we can summarize the position and relative importance of each node at a particular time. We call these geography-specific compression summaries \textit{geographic signatures} (Section~\ref{geosigs}). 

Using low-dimensional representations of graphs has been effective in network analysis tasks \cite{cao2015grarep, tang2015line, dai2018adversarial}. Additionally, graph neural network (GNN)-based approaches have been successful in generalizing outside of trained graphs \citep{zhou2020graph}, although it is not clear how specific link-recommendations would be ordered to avoid over-recommending. We want to use the geographic signatures to capture historical network structure and infer relative importance between origin nodes and a destination (a rank), while also allowing experimentation with future network structure. Using geographic signatures along with live operator-feedback, we show that we can learn the relationship between historical network activity and node pair flow/connection. Although rank features have been used in TSP problems \cite{vasko2011statistical}, it has not been treated as a target that is updated and reinforced. Online learning approaches via low-dimensional representations have been looked at for social network learning \cite{perozzi2014deepwalk}. Here, we develop methodology to extend the idea to operations bound by geography. In Section~\ref{networks_rl}, we propose a framework for using the geographic signatures to train a probabilistic model that ranks potential future connections and learns based on actual user decisions. We conclude with future work. 

\section{Summarizing Transportation Network Activity Using Geographic Signatures}\label{geosigs}

In this paper, we describe our work on the transportation problem of an e-commerce company, referred to here as $XYZ$. At $XYZ$, inventory is stored within warehouses across the United States (Fulfillment Centers or FCs for short). From within the FC, a purchased item from the website is picked, packed and moved onto a truck to be moved through the transportation network. A package at an FC can move to a sortation node where a package will get sorted and put onto a truck with other packages going to a common delivery node. Alternatively, a package can also move directly to a delivery node without going through a sortation node. There are other variations possible, but the transportation of a package between the origin warehouse and a delivery node is what is known commonly in the industry as the 'middle mile.'  

A central problem in middle-mile network design is predicting origin-destination (OD) flows: How many packages will flow from each origin warehouse to each Zip3/Zip5/facility/etc on a given date in the future? There are many possible approaches to this problem across the industry: optimization via mixed integer or linear programming or forecasting via statistical time series models are two major examples. However, in the optimization case, it is easy for the number of variables to blow up for a problem that has a large network or a large horizon of time, so simplified problems are typically considered that preserve structure and process as much as computation allows. While in the statistical case, it can be difficult to effectively capture geography and other process constraints, but time and uncertainty can be modeled readily. We want to focus on capturing geography in a problem. 

In this paper, we introduce a mathematical construct for bringing in geographical information into modeling, by using a concept we call geosigs (abbreviation for “geographic signatures”). This concept’s construction draws from the field of lossy image compression \cite{al1998lossy}. The core concept is to leverage spatial structure to achieve significant compression. In this paper, we focus on a well-structured network for transportation, but Appendix A shows how we might build a Flappy Bird agent using compressed images, where Flappy Bird has little structure in comparison to a transportation graph. After motivating the transportation problem in Section~\ref{geosigs_problem}, we define the compression summary for this problem as geographic signatures (\ref{geosigs_definition}). We demonstrate the value of geographic signatures by building a simple OD flow prediction model for the United States and observe how the geographic infromation can be utilized for prediction and experimentation. 

\subsection{Motivating Problem}\label{geosigs_problem}

\begin{figure}[ht]
	\centering
		\vspace{-0.1in}
	\includegraphics[height=1in,width=3in]{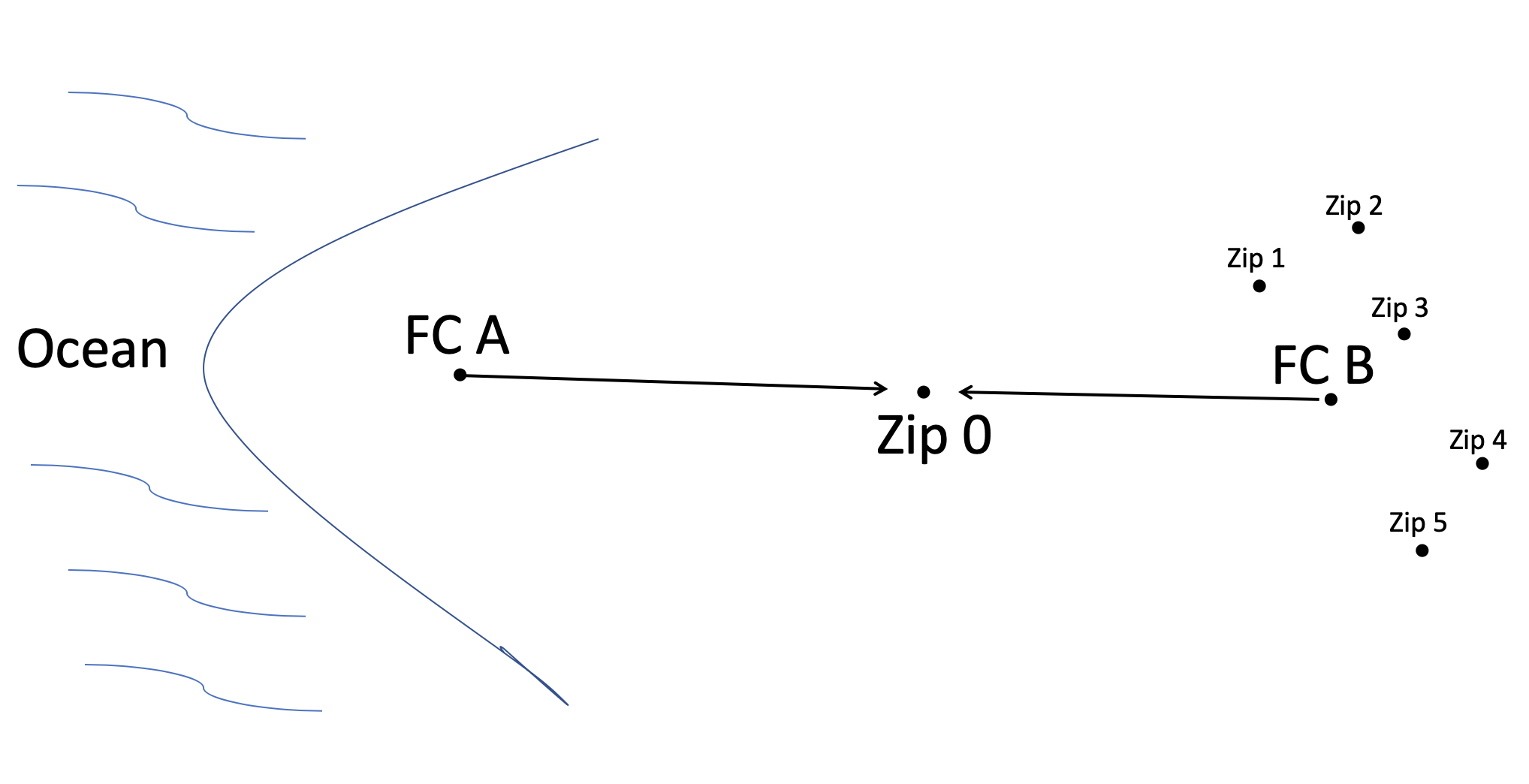}
	\caption{Mock topology. }
	\label{fig:geosig_problem}
		\vspace{-0.2in}
\end{figure}

Consider the situation in Figure~\ref{fig:geosig_problem}, where we have two origins at equal distance from destination $Zip_{0}$, where Zip0 is some zip-code of customers. The question we are trying to answer is the following: “How much volume flows from each FC to Zip0?” A purely distance-based model will assign equal volumes to both FCs, because they are equidistant from Zip0.

However, the picture shows some potentially useful geographic information. $FC_A$ is surrounded by an ocean, with no population centers around it. $FC_B$, on the other hand, is surrounded by population centers. One would therefore expect that $FC_B$ sends much of its volume to its nearby population centers, and has very little to send to $Zip_0$, while $FC_A$ fulfills a disproportionately higher fraction of $Zip_0$’s demand. It is possible to capture this implicitly via capacity constraints, but our premise is that we should not need to go to force the constraint.We should enable our models to capture what is easily visible to the human eye: the different geographic embeddings of the two FCs must result in different predicted flows to $Zip_0$. The rest of this section describes how we propose doing this, and what our test results show.

Throughout, we use historical flow data from $XYZ$ that has been masked. All analyses are done using Python on an i7 2.5 Ghz machine with 16 GB RAM.

\subsection{Geographic Signatures Definition}\label{geosigs_definition}

We begin by providing the formal definition of how we compute compression summaries in this setting, followed by detailed discussion of each step. For notation's sake, let $ZipN$ denote the first $N$ digits of a zip code, e.g. for 98109-4385, Zip3=981, Zip5=98109, and Zip9=98109-4385. Also, we abbreviate latitude (`lat') and longitude (`lon').

Let $G$ be a collection of nodes, such as the collection of all Zip3s, or the collection of all FCs. Let $x$ be a single node, which may or may not belong to $G$. For every node, we are given the latitude, longitude, and a measure $f$. The measure may be just a constant 1 in the simplest case; or it may be something more appropriate for the context: e.g. if $G$ is the collection of all FCs, then $f$ may be the capacity of each FC, or if $G$ is the collection of all Zip3s, it may be the “purchasing power” of each Zip3. 

The compression summary, $CS(\cdot,\cdot)$ of $x\in G$ is defined as:
\[
CS(x,g) = PrincipalComponents(FFT(Polar(x,G))).
\]

{\bf $Polar(x,G)$}: This is a matrix that represents $(G,f)$, as viewed from $x$, in polar coordinates, with some discretization. Consider a node $v$, defined by $(lat(v),lon(v),f(v))$. The polar representation of its location with respect to $x$ is given by $Polar(x,G) = (r,\theta)(x,v)$, where $r =   \sqrt{\left( lat(v)-lat(x) \right)^2 + \left( lon(v)-lon(x) \right)^2 }$ and $\theta = \tan^{-1}\left(\frac{lat(v)-lat(x)}{lon(v)-lon(x)} \right)$, with appropriate manipulation of the sign of $\theta$ so that starting from East and rotating counter-clockwise, $\theta$ goes from 0 to $2\pi$ (in radians).

We choose the step sizes of $r$ and $\theta$. For $\theta$, suppose we can choose the four cardinal directions, steps of $\frac{\pi}{2}$, or smaller steps of $\frac{\pi}{16}$. For $r$, choosing $r=1$ on a map in latitude and longitude is similar to choosing $r'\approx69$ miles. We might choose $r$ as the expected length of a truck route to reflect the actual segmentation of a transportation network. For any chosen set of step sizes, we sum the $f$ values for all nodes in $G$ that fall into a common neighborhood of $r$ and $\theta$. In general, applying polar coordinates with discretization yields $P\in\mathbb{R}^{d_r\times d_\theta}$, where $d_r$ and $d_\theta$ denote the number of discrete points in the range of $r$ and $\theta$, respectively.

\begin{figure}[t]
	\centering
	\subfloat[ Los Angeles ]{\includegraphics[height=1in,width=1.5 in]{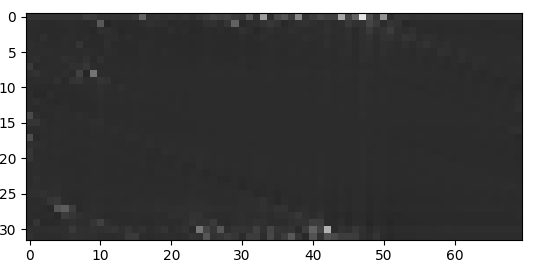}}
	\subfloat[ Houston ]{\includegraphics[height=1in,width=1.5 in]{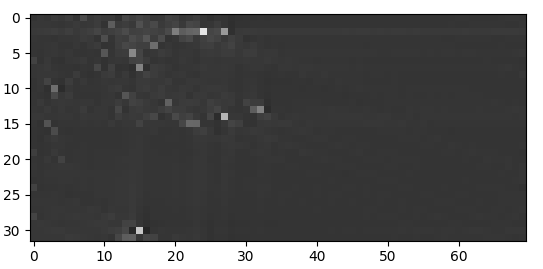}} 
	\caption{Polar matrices (as bitmaps) representing flows from two FCs, in Los Angeles and Houston, to a collection of Zip3s for a particular day. In this example, $r\in[0,70]$ ($x$-axis), $\theta\in[0,30]$ ($y$-axis descending), and $P\in\mathbb{R}^{d_{70}\times d_{30}}$. }
	\label{fig:la}
		\vspace{-0.2in}
\end{figure}

We can visualize this polar matrix as a bitmap. Two such bitmaps are shown below for two FCs in the $XYZ$ transportation network (Figure~\ref{fig:la}), one in Los Angeles and the other in Houston (each FC can be treated as the top left corner). In each, $\theta$ is on the Y-axis, with $Y=0$ representing East, and as we go down the Y-axis, going to North, West, South, and back East. The X-axis represents distance. Here, $G$ is the set of all Zip3s, and $f$ is the total volume at each Zip3. We see that a lot of the far distances (the right half of the images) are empty and this is mainly because an FC in LA may not typically be the origin of packages going into the eastern corners of the US. A little more than just the bottom half of the LA picture is also dark (zero measure), which is explained by the presence of an ocean Northwest, West, and Southwest of LA. Houston has some destinations West and close, but not far. 

{\bf $FFT(Polar(x,G))$}: Next we apply a Fast Fourier Transform to the matrix $Polar(x,G)$ yielding a matrix $F\in \mathbb{C}^{d_r \times d_\theta}$. FFT is also an invertible matrix transformation, from the real space to the complex space with the same dimensions. It decomposes the original image using simpler basis functions. A consequence is information “concentration”: FFT (like other such transforms) forces most of the signal to be at matrix positions closest to (0,0) (as we have defined it).

{\bf $PrincipalComponents(FFT(Polar(x,G)))$ }: Finally, taking advantage of the information concentration done in the FFT step, we select the entries $(i,j)$ of $F$ such that $i+j\le mask_{max}$, which we call the principal components of the matrix for ease of exposition. In one of the results displayed, we only selected 3 matrix ($mask_{max} = 1$) elements to summarize the position between each OD pair: (0,0), (0,1) and (1,0). This is where we accomplish compression: the entire picture is represented by 3 complex numbers, or equivalently, 6 real-valued numbers. In fact, the imaginary component of the (0,0) element of the FFT is always 0, so only 5 numbers represent the entire picture. These 5 numbers define the compression summary, or $CS(x,G)$.

\begin{figure}[t]
	\centering
	\subfloat[ $mask_{max}=1$ ]{\includegraphics[height=1in,width=1.5 in]{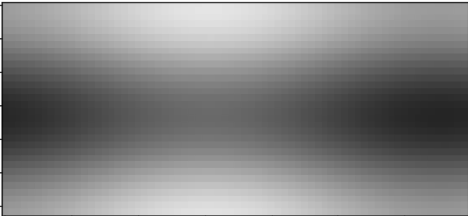}}
	\subfloat[ $mask_{max}=3$ ]{\includegraphics[height=1in,width=1.5 in]{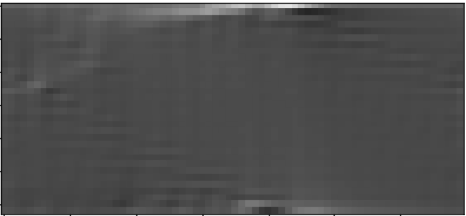}} 
	\caption{For Los Angeles, the image after applying a mask on $F$ and inverting onto the magnitude spectrum.}
	\label{fig:la_mag}
	\vspace{-0.3in}
\end{figure}

In Figure~\ref{fig:la_mag}, we show two pictures which are inverses of the LA picture displayed on the previous page, with $mask_{max}$ equal to 1 and 3 respectively. Observe that neither picture is exactly equal to the LA picture shown in Figure~\ref{fig:la}, e.g. “lossy” image compression. Selecting more principal components (higher $mask_{max}$) results in an inverted image that is close to the original: selecting a smaller value of $mask_{max}$ results in greater information loss. In that sense, $mask_{max}$ may be thought of as a “tuning parameter.”

In terms of measure, these summaries do not reside on the domain of the original image, which is why in the previous example we had to invert the image. In the original images on the real domain of flows, each row/column pair correspond to an angle/distance historical flow. Once we apply the mask on the coefficients and transform, we get instead a smoothing of flows on a magnitude spectrum. In signal processing, each complex coefficient $\alpha^*=(r,c)$ can be transformed such that the power, $power$, can be expressed as $power=(r^2+c^2 )^{(1/2)}$. Here, that power is interpreted as instantaneous flow or flow intensity.

When we use the first three upper left corner coefficients using $mask_{max}=1$, then we get the average flow intensity at the origin (0,0), 1 degree away from the origin (1,0), and 1 turn or angular step from the origin (0,1). For the LA image, this might be useful for northern flow activity, but it will not capture  southern flow activity. With a mask of 3, we increase the number of coefficient and flow intensity estimates to 6 (so 6 potential features for a regression model). However, we learn nothing about where flows occur, only about the lack of flows close to the FC, which is useful information for problems dependent on local network activity. Ideally, the level of discretization is fully utilized, since this is the most preserved level of information. In Figure~\ref{fig:la_mag}, we observe how discretization can be interpreted on the magnitude spectrum (inverted image) as a smoothing of the observed flows. We propose a summary of the magnitude spectrum that fully utilizes the discretization in a meaningful way for many network problems. 

The geographic signature, $Geosig(\cdot,\cdot)$ of $x\in G$ is defined as:
\[
Geosig(x,g) = MagnitudeSpectrum(CS(x,G)).
\]
We use this transformation to summarize network structure and reduce dimensionality of the feature space. We propose to discretize the angles of the network into 4 angles: $(0,\pi/2),(\pi/2,\pi),(\pi,3\pi/2)$,and $(3\pi/2,2\pi)$. This discretization follows the 4 directions: NE, NW, SW, and SE (in that order). Next, for the radius of the network, we choose 100 mile segments, which with 17 segments in total would cover most of the continental United States of America. In Section~\ref{geosigs_definition}, we left $r=1$ at the default of 1 degree (about 69 miles per segment), switching to 100 mile segments instead reduces the dimensionality for the subsequent examples, but in practice this would need to be chosen to represent the minimum segment distance of importance to the network. 

Hence, we discretize each network flow image into 4 rows and 17 columns. We propose using a semi-coarse mask ($mask_{max}=2$ to be discussed) and summarizing the $4*17=68$ geosigs by the maximum smoothed flow along each row or direction. Furthermore, each max flow is paired with the position of the max, i.e. the number of 100 mile segments that it takes to get to that maximum flow intensity. In this way, we preserve specific and interpretable information, namely, how far a Zip3 can be from a particular FC to be fulfilled.

The geographic signatures or geosigs can be tailored to the problem and extract key metrics. This customization is especially important if a network is multilevel, where different granularities (Zip2, Zip3, Zip5, Zip9, Mean Statistical Area (MSA), etc.) play different roles in strategic decisions, but they also can be combined. Using geosigs, we get dimensionality and interpretability benefits, because we compressed along an image specified with structure (selected increments/angles) and extracted the problem-specific metrics. This compression and summary approach is applicable in any spatial problem, see Section~\ref{flappy_inputs} for a Flappy Bird example with minimal spatial structure. For other transportation problems, alternatives for geosig metrics include using the first non-zero flow intensity (requires a fine mask) or using the first flow intensity above an interesting threshold (say 3000 packages). In what follows, we demonstrate the utility of the first proposed summary of geosigs, the max flow pairs. 

\subsection{Explaining Historical OD Flow with Geographic Signatures}\label{geosigs_maxflowpairs}

We now demonstrate the value of compression summaries and geosigs by building a simple OD flow prediction model. In practice, OD flow prediction would not be handled by such a simple model, but the model simplicity allows us to easily track the information contribution of the geosigs.

The problem statement is the following: for a single selected date in the future, predict exactly how many packages will flow from each FC to each Zip3. If items were in inventory at the closest facility to a customer always, then this would not be such a difficult problem. Instead, we have to expect that many facilities will help to fulfill the demand from a particular destination, either because of rarely ordered items or items that can only be stored in special FCs. In this section, we estimate a linear regression model where the dependent variable is the number of packages in each $FC\rightarrow Zip3$ pair, and we use several independent variables (geosig related) explained next. Note that had we included other non-geosig related features, we would want to use a more complex model to correctly share information across features (as in Section 3.1). Models considered are trained on a single day: this is day 221 in the year 2018, which is Thursday, August 9, 2018 (18,635 pairs).  We test the models against day 249 of 2018, which is Thursday, September 5, 2018 (21,042 pairs). The flow data used in these studies has a controlled amount of noise to retain a particular degree of the original signal, while aliasing the true values of the confidential data. We evaluate the utility of the geosigs as explanatory and predictive variables using simple regression models evaluated via mean actual percentage error (MAPE) and adjusted $R^2$. At the scale $R^2$ is measured, we introduced at least 80\% of the variation in the noisy data, so $R^2>0.10$ will be desirable.

The linear regression models use the following independent variables:
\begin{itemize}
	\item ln\_lld: Natural logarithm of the Euclidean lat-lon distance between the FC and the Zip3. Included in all models.
\end{itemize}
Model A includes 14 geosig variables (dDI\_01, dDI\_10, dDR\_00, dDR\_01, dDR\_10, oOI\_01, oOI\_10, oOR\_01, oOR\_10, oDI\_01, oDI\_10, oDR\_01, oDR\_10, and oDR\_00), defined according to the following convention:
\begin{itemize}
	\item oDR\_00: This computes the geosig of each origin FC, with respect to the matrix of all FCs, and extracts the real part at matrix location (0,0).
	\item dDI\_01: Geosig of each Zip3, with respect to the matrix of all Zip3s, imaginary component, matrix location (0,1).
\end{itemize}
Model B uses magnitude or power of the first coordinate (interpreted as average flow intensity) of each domain, denoted as oO\_00, oD\_00, and dD\_00. Model C and D use 4 max flow distance pair values (8 values per OD pair), with the following convention:
\begin{itemize}
	\item oD\_0r\_summary\_max: This computes the max intensity along the 0th direction $(0,\pi/2)$ for $FC\rightarrow Zip3$. 
	\item oO\_0r\_summary\_max: This computes the max intensity along the 0th direction $(0,\pi/2)$ for $FC\rightarrow FC$. 
	\item dD\_0r\_summary\_max: This computes the max intensity along the 0th direction $(0,\pi/2)$ for $Zip3\rightarrow Zip3$. 
	\item oD\_0r\_summary\_max\_r: This computes the distance, $r$, at which the max intensity occurs along the 0th direction $(0,\pi/2)$ for $FC\rightarrow Zip3$. 
\end{itemize} 
Model C uses only the max flow (4 values per OD pair). These are all computed with the measure $f$ being the total flow on the training date: that is, for the matrix $D$, the measure $f$ of each Zip3 is the total number of packages delivered to that Zip3, and likewise for FCs in O. We use three sets of geosigs: origin domain (o,O),origin-destination domain (o,D), and destination domain (d,D). For each domain set, we extract the real components at matrix indices (0,0), (0,1) and (1,0), and the imaginary components at (0,1) and (1,0).

In Table~\ref{table:rsquares}, we present the results of fitting linear models using geosigs. The best predictor of day 249 was Model B with 57\% MAPE, although it had a low $R^2$. Using the average flow intensity along each domain resulted in a low variance, but higher bias estimator. We see that raw geosigs can explain about 19\% of the variation in the training data (Model A), but summarizing the geosigs not only improves test MAPE (Model C, Model D), but also provides a more generalizable model (Model D with highest $R^2$). 

\begin{table}[t]
	\centering
	\small
	\begin{tabular}{c|ccc}
		\hline
		Model & Adjusted $R^2$ & Train-MAPE & Test-MAPE \\ 
		\hline
		Null & 0.01 & 0.96 & 0.87 \\
		A & 0.19 & 0.87 & 0.79 \\
		B & 0.06 & 0.87 & 0.57 \\
		C & 0.13 & 0.85 & 0.78 \\
		D & 0.31 & 0.77 & 0.71 \\
		\hline
	\end{tabular}
	\caption{Adjusted $R^2$ values for linear models fit on flow data using features engineered using geographic summaries. We also include results for the null model which excludes any geosig related features (Null). } 
	\label{table:rsquares}
		\vspace{-0.3in}
\end{table}

Overall, this shows the improvements we gain by tailoring geographic signatures for a flow problem. By using geographic signatures, we reduced the dimensionality of the network, preserved the signal relevant to the relationship between geography and OD flow, and retained features that are interpretable. We next show how to use that interpretability to learn fundamental network properties that allow us to consider hypothetical networks. 

\subsection{Hypothetical Networks}\label{geosigs_hypothetical}

All of the measures that we are considering as belonging to a network or topology are intrinsically dependent on each other. This means even when we summarize the information into geographic signatures there is correlation among the features. We want to capture this collinearity to have an understanding of network effects that generalizes to other within-network scenarios. We fit a gradient boosted random forest to capture some of that dependence and demonstrate how the geosigs allow us to play out basic scenarios such as arc additions or new facilities.  

\begin{figure}[t]
	\centering
	\includegraphics[height=1.25in,width=3in]{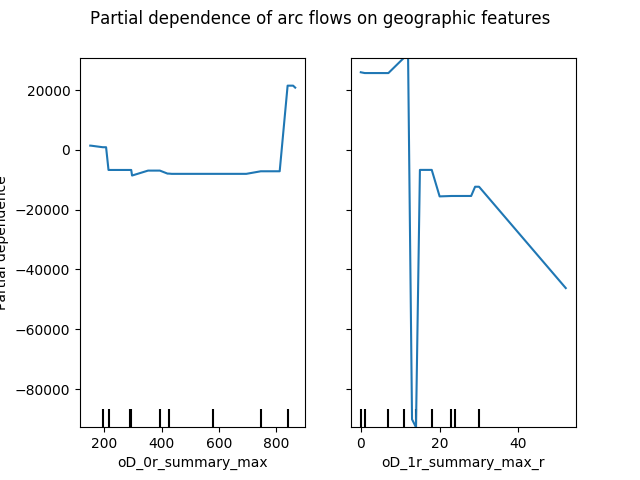}
	\caption{PDPs look at a variable of interest across a range. At each value, the model is evaluated for all observations of the other model inputs, and the output is then averaged. A negative (positive) value indicates that the feature value may be associated with smaller (greater) flows compared to the rest of the range averaged across the other features. The ticks mark quantiles.}
	\label{fig:hypo_train}
		\vspace{-0.2in}
\end{figure}

Using the features in Model D (distance and the max flow pairs), a gradient boosted regression forest was trained (\texttt{sklearn} with 1000 boosting iterations, otherwise default settings) and had a test MAPE of 63.6\%. When observing the flow tendencies of the network between the strongest pairs of FCs and destinations in the NE direction ('oD\_0r\_summary\_max'), the pairs with an average flow intensity less than 800 boxes (about 90\% of arcs based on Figure~\ref{fig:hypo_train}) tended to have a similar overall flow that was smaller than the flow expected from pairs with an average flow intensity greater than 800 boxes. These NE-directed max pairs tended to occur quite far (close to 60\% occurred at more than 1500 miles) and those large distances tended to be associated with smaller flows overall as can be seen in the PDP for the distance of the max pair, 'oD\_0r\_summary\_max\_r'. Overall, this suggests a network that has a particular relationship with the NE direction: it has FCs that serve it thick arcs from about 1500 miles, but for the most part NE directed arcs are not that thick. 

Using the partial dependence values directly, we can estimate the contribution of a particular type of activity to a network. In the example given, we can estimate the total change in flow to a network relative to the activity of NE-moving packages. Under the network represented by the data sampled and transformed, we can expect 8931 fewer boxes moving NE than the average of the other directions. NW moving boxes are the minority averaging about 10,253 fewer boxes than the network average. The most active overall direction under this summary is the SW set of boxes at about 6491 above average. 

Suppose we place a hub in the middle between Los Angeles and New York at about 800 miles from Los Angeles, so that we can reduce the overall average arc length of the network, call it $HUB$. Using the PDPs, we estimate changes in flow across the network. We want $HUB$ to play an important role for the SW region, so we want network information to reflect this new activity. We generate data such that all origins SW of $HUB$ will have 'oD\_0r\_summary\_max\_r' equal to the number of segments between the origin and $HUB$. 'oD\_0r\_summary\_max' might tend to be high around 800 boxes. Other than this behavior change in the network for the SW region, $HUB$ is not going to impact any other origins. This implies that getting the average value for all other geosigs will suffice. 

In a small experiment, we replace 25\% of the test data to reflect the SW region's new $HUB$ and randomly generate distance and two of the geosigs as Gaussian random variables centered at the means of interest. Partial dependence on generated data quantifies how the network tends to be associated with that activity summary. Using the geosigs for the historical network, we can expect 17,382 fewer packages going NE with the addition of $HUB$, but 12,226 more boxes to be moving SW. When one area of the network experiences more activity, the learned network associates that with a decrease in activity for other areas. This experiment highlights the dependence that can be captured using geographic summaries. Under this experiment, we would need to know specific arc cost information to estimate whether this $HUB$ would lead to overall reduction in network costs or other important metrics. In the next section, we consider the problem of network activity at a larger scale and utilize the geographic signatures to capture network structure. 

\section{Lane Recommendations Reinforced by Users}\label{networks_rl}

We define the short term problem of deciding arcs for the next time period based on last time period as a cascading bandit problem \cite{russo2018tutorial}. We train a probabilistic model, a cascade model, that governs the arcs that will be recommended to a business operator. This probabilistic model can be informed by our prior knowledge of the network as encoded by the geographic signatures. In Section~\ref{networks_ranks}, we rank arcs particular to a destination based on cost of a package on that arc derived from a regression model that takes as features geographic signatures among others. We rank by cost because we want to minimize the overall cost of the network. Using these ranks as priors, we propose a Thompson sampling algorithm to learn the probabilistic model that governs arc recommendations (Section~\ref{networks_ts}). Using Bayesian techniques for reinforcement learning, we show in Section~\ref{networks_results} that it does not take many iterations to learn important network structure under a consistent network topology. We also discuss user-driven agent-assisted experiments to summarize new topologies.

\subsection{Ranking Arcs Using Geographic Signatures}\label{networks_ranks}

Using historical arc costs (based on truck costs) and volumes, we want to predict future arc connections. To do this, we will rank the arcs based on cost and rank low cost arcs high. First, we must estimate cost. For demonstration, we propose a regression model for weekly data with the following data structure. Let $FC_i$ and $D_j$ denote the $i$th origin FC and the $j$th destination node. $D_j$ can be a delivery node or a hub. We define $c_{i,j,t}$ as the average historical cost per package and $n_{i,j,t}$ as the number of packages or flow on arc $FC_i\rightarrow D_j$ at time $t$. Additional measures we consider in this demonstration include: $FC_i$ latitude and longitude, $lat_i$ and $lon_i$; the distance between $FC_i$ and $D_j$, $d_{i,j}$; and a binary indicator $direct_{i,j}=1$ when $D_j$ is a delivery node (otherwise, 0).  

We need to be able to capture network topology of the United States. We want to know when a facility serves locally more than across the US or when there is a body of water or a mountain that explains why a facility does not serve a destination. However, we want to embed this information as simply as possible. Using flow and location $(f_{i,j}, lat_i, lon_i, lat_j, lon_j)$, we define geographic signatures as done in Section~\ref{geosigs_definition}. In particular, we use the max flow pair summaries introduced in Section~\ref{geosigs_maxflowpairs}, where 
$g_{i,j}=geosigs(f_{i,j}, lat_i, lon_i, lat_j, lon_j)\in\mathbb{R}^{8}$. Let $x_{i,j,t}=[t, direct_{i,j}, f_{i,j}, d_{i,j}, lat_i, lon_i, g_{i,j}]$. 
We aim to estimate the expected cost of an arc at time t, 
\[
\chat_{i,j,t}=\E \left[ c_{i,j,t} \mid x_{i,j,t} \right]= \fhat(x_{i,j,t})
\], 
where $\fhat$ can be estimated using random forest regression techniques.

Using 52 weeks of historical data (spanning October 2018 to October 2019), we fit daily $\hat{f}_{i,j,t}$ to predict $\hat{c}_{i,j,t'}$ ($t'>0$). We pre-process the arcs to retain arcs with a minimal flow and met other confidential requirements. We consider 71 origin FCs and 232 destinations. For each $D_j$, let $sorted_{j,t'}=(FC_{i',j}, FC_{i'+1,j},\ldots)$, where $\hat{c}_{i',j,t'}\le \hat{c}_{i'+1,j,t'} \le \ldots \le \hat{c}_{max,j,t'}$. The position of $FC_i$ in $sorted_{j,t'}$ denotes the rank relative to $D_j$, $rank_{i,j,t'}=index(sorted_{j,t'}==FC_i)\in[0,1,2,..., 70]$. 

We see in Table~\ref{table:d0_ranks} the rankings for $D_0$ alongside distance for two weeks, week 45 and 52. Of note, the ranks do not obey a distance law. In Table~\ref{table:d0_actuals_nonconfidential}, we show a glance of the top 10 week 52 actuals (see Table~\ref{table:d0_connections} for the number of connections per week). We see that the top 10 estimated ranks are similar to the top 10 actuals, except $FC_{56}$. New connections open as the peak season gets closer, so we expect peak-specific arcs to appear. Since the estimated ranks are time-dependent, then we expect some increase in ranks of infrequent arcs, but if a destination has a consistent set of 10 or more then we may not observe it in a top 10 rule. To take advantage of the recurring cycle of decisions and improve the chance we observe infrequent but necessary arcs, we propose a probabilistic model that updates based on user decisions using Thompson sampling as discussed next. 

\begin{table*}[t]
	\centering
	\small
	\begin{tabular}{c|cc|cc}
		Rank & Week 45 FCs & Distance & Week 52 FCs  & Distance \\ 
		\hline
		1 & 1 & 14.55 & 1 &  14.55 \\
		\hline
		2 & 7 & 115.04 & 52 & 48.01 \\
		\hline
		3 & 52 & 48.01 & 7 & 115.04  \\
		\hline
		4 & 2 & 323.77 & 35 & 119.21  \\
		\hline
		5 & 14 & 110.57 & 21 & 83.53 \\
		\hline
		6 & 35 & 119.21 & 0 & 89.32 \\
		\hline
		7 & 0 & 89.32 & 38 & 1378.55 \\
		\hline
		8 & 21 & 83.53 & 31 & 137.60 \\
		\hline
		9 & 31 & 137.60 & 2 & 323.77 \\
		\hline
		10 & 38 & 1378.55 & 14 & 110.57 \\
		\hline
	\end{tabular}
	\caption{Ranks for $D_0$. Rank 1 corresponds to the arc between an FC and $D_0$ with the lowest expected cost per package. } 
	\label{table:d0_ranks}
\end{table*}

\begin{table}[t]
	\centering
	\small
	\begin{tabular}{c|cc|cc}
		Rank & Week 45 FCs & Trucks & Week 52 FCs  & Trucks \\ 
		\hline
		1 & 1 & 63 & 20 & 27  \\
		\hline
		2 & 14 & 60 & 11 & 26 \\
		\hline
		3 & 7 & 53 & 1 &  25 \\
		\hline
		4 & 35 & 51 &  31 & 22  \\
		\hline
		5 & 31 & 47 &  14 & 20 \\
		\hline
		6 & 21 & 46 &  21 & 19 \\
		\hline
		7 & 0 & 27 &  12 & 18 \\
		\hline
		8 & 52 & 24 & 56 & 17 \\
		\hline
		9 & 12 & 22 & 0 & 17 \\
		\hline
		10 & 11 & 19 & 52 & 15 \\
		\hline
	\end{tabular}
	\caption{Actuals for $D_0$. Rank 1 corresponds to the arc between an FC and $D_0$ with the largest number of trucks (we show only top 10). We only show actuals for the FCs considered in this study. } 
	\label{table:d0_actuals_nonconfidential}
			\vspace{-0.1in}
\end{table}

\begin{table}[t]
	\centering
	\small
	\begin{tabular}{c|c}
		Week & Number of Connections \\
		\hline
		45 & 17 \\
		46 & 22 \\
		47 & 18 \\
		48 & 18 \\
		49 & 19 \\
		50 & 16 \\
		51 & 22 \\
		52 & 18 \\
		\hline 
	\end{tabular}
	\caption{Number of actual connections between the study FCs and $D_0$. Mean=18.75, Standard Deviation=2.19.}
	\label{table:d0_connections}
\end{table}

\subsection{Thompson Sampling to Probabilistically Learn Future Arcs}\label{networks_ts}

The cascading bandit model for this problem is defined by $(N_D, N_{FC},K_j,\theta)$, where $N_D$ is the number of destinations, $N_{FC}$ is the number of FCs, $K_j\le N_{FC}$ is the number of arcs recommended at time $t$ for $D_j$, and $\theta\in[0,1]^{N_{FC}}$ is a vector of connection probabilities. In a cascading scheme at time $t$, an agent recommends an ordered list $a_t\in\{1,\ldots,N_{FC}\}^{K_j}$. The user selects $k\le N_{FC}$, call these selected arcs $a'_t$. 

For arc $FC_i\rightarrow D_j$ denoted by $arc_{i,j}$, let $\alpha_{i,j}\equiv$the number of times the arc was selected and $\beta_{i,j}\equiv$the number of times the arc was not chosen, but was observed in the top $K$. Finally, to use the rankings as a weight for future decisions, let $rankpct_{i,j}=1-\frac{rank_{i,j}+1}{N_{FC}}$. In Algorithm 1, we describe the algorithm for using prior information to update the probability of selecting an arc, $arc_{i,j}$ at time $t$. Note that the initial probabilities of connections are sampled independently (a false assumption for network connections). However, we estimated ranks based on costs that have accounted for dependencies at some high level. Assuming overall structure of a network does not change significantly within short time intervals, then the geographic signatures presents a useful structured ranking. The updated connection probability reflects that information if it is reinforced by the user enough times. We show that this algorithm captures structure in a timely manner. 

\begin{algorithm}[h]
	\SetAlgoLined
	\For{$t=1,2,\ldots$}{
		\# Sample probability of connection; \\
		\For{$j=1,2,\ldots,N_{D}$}{
			\For{$i=1,2,\ldots,N_{FC}$}{
				Sample: $\hat{\theta}_{i,j}\sim Beta(\alpha_{i,j}, \beta_{i,j})$. \\
				Adjust $\tilde{\theta}_{i,j}=\hat{\theta}_{i,j}rankpct_{i,j}$.
			}
		}
		$a_t \leftarrow$ arcs with $K$ largest $\tilde{\theta}_{i,j}$.
	}
	\# Update posterior distribution hyperparameters; \\
	\For{$t=1,2,\ldots$}{
		\# Sample probability of connection;\\
		\For{$j=1,2,\ldots,N_{D}$}{
			\For{$i=1,2,\ldots,N_{FC}$}{
				$\alpha_{i,j}\leftarrow  \alpha_{i,j} + \mathbf{1}\{arc_{i,j}\in a'_t \}$. \\
				$ \beta_{i,j}\leftarrow  \beta_{i,j} + \mathbf{1}\{ arc_{i,j}\in a_t \}$.
			}
		}
	}
	\caption{$ArcsTS(N_{FC},K, \alpha, \beta)$}
\end{algorithm}

\subsection{Experimental Results}\label{networks_results}

\begin{figure}[h]
	\centering
	\includegraphics[height=1.5in,width=3in]{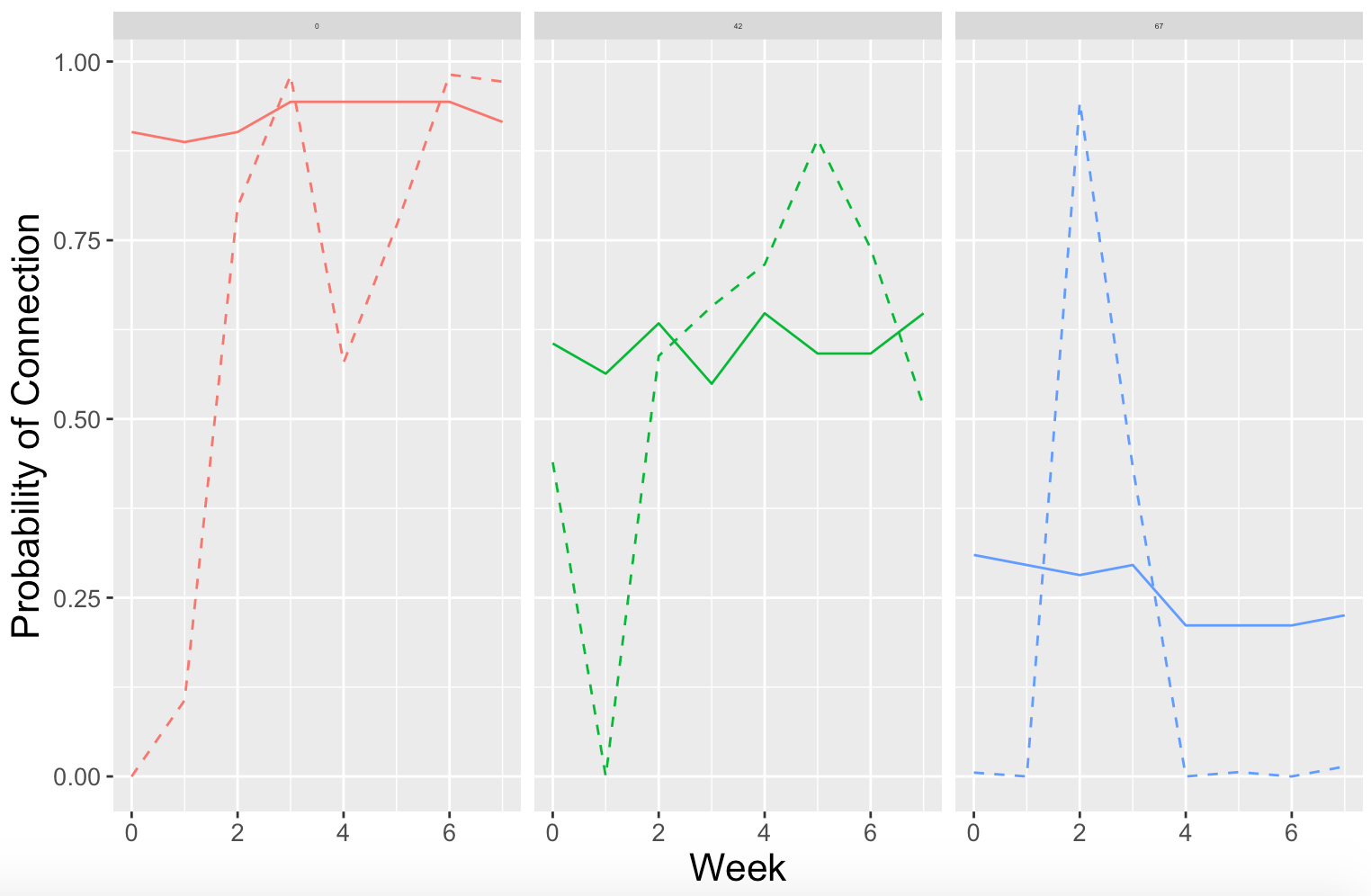}
	\caption{Rank percentile (solid) and connection probabilities (dashed) with $D_0$ across 8 weeks for 3 random FCs (0, 42, 67).  }
	\label{fig:d0_learning}
		\vspace{-0.1in}
\end{figure}

Using 8 weeks of data, we run Algorithm 1 to recommend 10 arcs. The probabilities can be initialized using $\alpha_{i,j}=0.1$ and $\beta_{i,j}=1.0$. Choosing an arc increases $\alpha$. If an arc is not chosen but in the top $K$ rule, then $\beta$ increases. For the initial week (week 45), the sampled probabilities of connection are uninformative, so on a first initialization we can use the ranks. In Figure~\ref{fig:d0_learning}, by the fourth week, FCs with rank percentiles greater 50\% start to get recommended with probability greater 50\%. FCs with rank percentiles below 50\% tended to be neither selected nor viewed, so by the 5th week had low probabilities (no rewards).

Table~\ref{table:d0_ts_results_confidential} shows the 10 arcs with largest sampled probability of connection to $D_0$. $F_{56}$ is recommended among the top 10. Although this is by probability, this case shows the usefulness of Bayesian reinforcement based on estimated ranking and operator decisions. $FC_{69}$ was not in the top 10 actuals for $D_0$, but it was active with only 2 trucks. 

\begin{table}[t]
	\centering
	\small
	\begin{tabular}{c|cc}
		Rank & Week 52 FCs  & Distance \\ 
		\hline
		1 & 52 & 48.01 \\
		\hline
		2 & 1 & 14.55   \\
		\hline
		3 & 7 & 115.04  \\
		\hline
		4 & 0 & 89.32  \\
		\hline
		5 & 35 & 119.21  \\
		\hline
		6 & 2 & 323.77   \\
		\hline
		7 & 22 & 133.57 \\
		\hline
		8 & 56 & 340.90  \\
		\hline
		9 & 69 & 1487.01  \\
		\hline
		10 & 14 & 110.57  \\
		\hline
	\end{tabular}
	\caption{Ranks for $D_0$. Rank 1 corresponds to the arc between an FC and $D_0$ with the largest sampled probability of connection. } 
	\label{table:d0_ts_results_confidential}
			\vspace{-0.3in}
\end{table}

One way to decrease the role of chance in the top $K$ rule (here $K=10$) is to introduce a top $K_j$ rule, where the number recommended depends on the $j$th destination. $D_0$ alone averaged about 19 connections (sd=2.19) over the 8 weeks. Other destinations play smaller or larger roles in terms of arc connectivity depending on time and network position. One approach to choosing $K_j$ is to use the probabilities of connection themselves. We can treat arc connectivity probabilities as belonging to a Bernoulli random variable, $Y_{i,j}\sim Bernoulli(\theta_{i,j}).$ Let $Y_j=\sum_i Y_{i,j}$. Assuming independence among $Y_{0,j}, \ldots, Y_{70,j}$, the expected number of arcs, $N^{arc}_j$, can be calculated as $N^{arc}_j=\mathbb{E}(Y_j)=\sum_i \theta_{i,j}$. In this way, we can use $\tilde{\theta}_{i,j}$ to estimate $\theta_{i,j}$ and say the agent wants to recommend $K_j=N^{arc}_j$ arcs. Under this formulation, we can also estimate the variability of the agent's recommendation total on destination $j$ as $\mathbf{Var}(Y_j)=\sum_i \theta_{i,j}(1-\theta_{i,j})$. For $D_0$ over the 8 weeks, the average $N^{arc}_j$ was 16 (sd=$\sqrt{\mathbf{Var}(Y_j)}=2.31$). During week 52, the agent would have recommended 24 arcs compared to the actual 18. 

Overall, this shows the agent is already converging on network behavior for $D_0$ as far as recommendations might be needed to. As a recommendation agent, we primarily want to reduce the search space for the next set of connections. We also want to be able to use the ranks and probabilities of connection to experiment with new networks. As discussed in Section~\ref{geosigs_hypothetical}, the geographic signatures can be used to summarize the fundamental properties of different networks. As shown here, these geographic signatures can be used to rank arc connections within a network. These ranks in turn can influence a cascading model's probability of connection to recommend a set of arcs. A simple framework for user-driven agent-assisted testing for the potential of a new network addition can be formulated as follows:
\begin{enumerate}
	\item Fit a model that captures important network dependencies using historical arc costs/flows data.
	\item Predict cost/flow on artificial data for new connections. 
	\item Use predictions to rank old and new connections. 
	\item Initialize new arcs using most similar existing arcs.
	\item Run the agent and see how close new arcs are to be included in the set $K_j$. With a range of artificial data sets, uncertainty of new arc utility can be measured. 
\end{enumerate}

There is more that can be done within Algorithm 1. For instance, a discount factor can be added so that connections that are guaranteed converge at some high probability. Moreover, it would be important to also introduce penalties for inactivity. One simple idea is to create a rule dependent on decreasing rank: lower $\beta$ by 1 if an arc drops in rank by some threshold. Finally, the most important addition to such an algorithm would be to capture the dependencies amongst the probabilities of connection in a way that still allows user-driven experiments. 

\section{Discussion}\label{discussion}

In this paper, we discussed image transformations for learning recurring decisions. We introduced summarizing complex geographic network structures using geographic summaries or geosigs. Geosigs captured fundamental network properties that can be helpful in ranking the potential of new connections. For recurring connections, we introduced a Bayesian reinforcement framework where an agent probabilistically determines connections based on ranks reflecting network value and user decisions. 

There is more that can be considered in future work both in terms of method and application. The probabilities and rankings can be merged into one probabilistic model that captures network dependencies more explicitly. Furthermore, tendencies in the agent's decisions can be used to track emerging networks in applications beyond transportation. 

\section{Declarations}

On behalf of all authors, the corresponding author states that there is no conflict of interest.

\section{Appendix}

\subsection{Playing Flappy Bird with Compression}\label{flappy}

\begin{figure}[h]
	\centering
	\includegraphics[height=1.5in,width=1.25in]{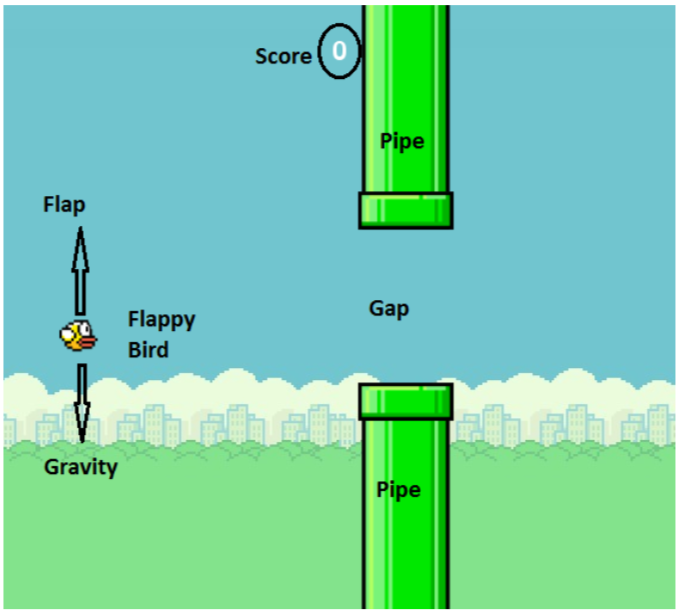}
	\caption{Flappy Bird schematic.}
	\label{fig:fb_schematic}
\end{figure}

In the paper, we focus on a transportation problem to demonstrate the utility of compression then extracting interpretable features from that compression for use in simple, but effective models for predicting flows and transportations decisions. In this section, we show for a problem with less spatial structure, i.e. the Flappy Bird game (Figure~\ref{fig:fb_schematic}), that compression is still a valuable addition by allowing us to offer only the relevant information in the image. This Flappy Bird example is complementary to the transportation problem presented in this paper, since we can discuss compressed structure of a game more candidly than a transportation graph loosely (due to noise) based on reality. We begin by discussing the image transformation and learning literature relative to games, then define how we use compression to execute the same learning task, i.e. train a successful Flappy Bird playing agent. 

Image transformations are used frequently in pre-processing for learning games. One of the first steps in developing a learner to play a game using sources such as PyGame or OpenAI's Gym (\cite{kelly2016basic}, \cite{nandy2018openai}) is to resize or denoise an image \citep{mnih2015human}. However, pre-processing can carry the risk of losing information. Resizing, for example, requires interpolation in order to maintain aspect ratio. Using deep neural networks to build up abstractions of the data, a game like Flappy Bird can be learned successfully using deep reinforcement learning from only mildly transformed pixel data (\cite{shu2014obstacles}, \cite{chen2015deep}, \cite{appiah2018playing}). Much of the machine learning literature that leans towards AI focuses on how algorithms can learn a task without the refinements of human knowledge and there has been remarkable success in this for complex games such as Go \citep{silver2017mastering}. The goal is to begin with little knowledge of what is important as perceived by humans, but allow the AI to learn via rewards and penalties earned through trial and error. 

In this section, we want to train a neural network to play Flappy Bird while pre-processing the screen images to reduce the dimensionality of the input as much as possible. To play, a user aims to move a bird between a pair of pipes by pressing a key to tell the bird to jump up, otherwise, the bird falls (Figure~\ref{fig:fb_schematic}). Not pressing a key allows the bird to lower in elevation, so it is necessary to adjust course for lower gaps. As a learning problem, we can take the game state to predict which action the bird should take: jump or not jump. One RL framework that has been particularly successful in tackling games such as Flappy Bird has been Deep Q-Networks (DQNs) as described in \cite{mnih2013playing} and \cite{mnih2015human}. In a DQN learned game, raw pixels are used as inputs to train successive convolutional neural networks (CNNs) with a variant of Q-learning. The output is a value function estimating future rewards. We build off of the PyTorch implementation described in \cite{piculjan2019}. 

We are interested in showing how pre-processing done to the raw pixels impacts training and testing performance. We begin by describing the different options we explore for pre-processing of the images, such as image compression (Section~\ref{flappy_inputs}). Then we describe the learning framework in which we will use those inputs in Section~\ref{flappy_learner}. Section~\ref{flappy_results} shows results for experiments where it can be seen that as the quality of information extracted from the raw pixels improved, so did the ability to earn rewards. For domain-agnostic learning in games this may not be too important, but for real industrial problems in a human environment with time between events of interest there is a lot to gain. 

\subsubsection{Pre-Processing of Flappy Bird Image}\label{flappy_inputs}

Using PyGame, Flappy Bird can be rendered and customized to the users liking. Whatever color or size preferences, the raw pixels are taken as the game state. From here, the basic formula introduced by~\cite{mnih2013playing} might include the following for DQN-learned Flappy Bird:
\begin{enumerate}
	\item Convert the pixels into gray scale.
	\item Resize the image.
	\item Stack last $K$ frames to produces an array of planes to be used by CNNs. 
\end{enumerate}
In (1), noise was removed, since the different colors do not matter: we only care about the bird and pipes. In (2), the number of pixels was reduced to limit the input dimension going into the neural networks. What is the smallest an image can go before it becomes uninformative? The last $K$ frames in (3) are used so that the transitions from one frame to another can be learned. One can imagine using fewer frames might lead to a bird that crashes often while actually between the pipes, while using more frames would lead to longer computation time. 

\begin{figure}[h]
	\centering
	\includegraphics[height=1.0in,width=3.1in]{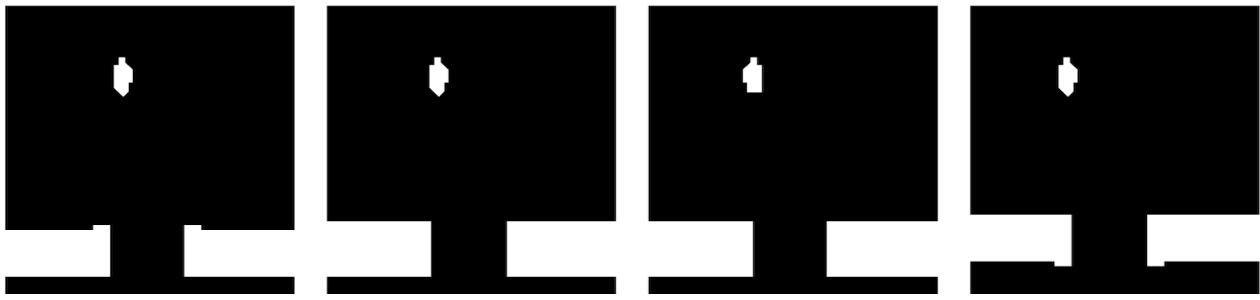}
	\caption{Set of four gray scale Flappy Bird game states resized to $84\times 84$ from~\cite{piculjan2019}. Note that these are rotated compared to Figure~\ref{fig:fb_schematic}.}
	\label{fig:fb_gray}
\end{figure}

In the experiments done in this section, we set $K=4$. We continue to convert to gray scale, but as in~\cite{piculjan2019}, we also take off the bottom of the screen to produce images such as Figure~\ref{fig:fb_gray}. 
The image size is originally $288\times404$, but can be resized down to $84\times 84$ as shown in Figure~\ref{fig:fb_gray}. We will experiment with resizing down smaller to $44\times 44$. 

\begin{figure}[h]
	\centering
	\subfloat[ ]{\includegraphics[height=1.0in,width=1.0 in]{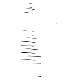}}
	
	\subfloat[ ]{\includegraphics[height=1.0in,width=1.0 in]{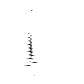}} 
	\subfloat[ ]{\includegraphics[height=1.0in,width=1.0 in]{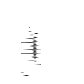}} 
	\subfloat[ ]{\includegraphics[height=1.0in,width=1.0 in]{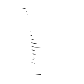}} 
	\caption{FFT of the images in Figure~\ref{fig:fb_gray}.}
	\label{fig:fb_hpf}
\end{figure}

We primarily experiment with image compression. After resizing and converting an image to gray scale, we get the Fast Fourier Transform (FFT) of the image, the zero frequency is shifted onto the position of the bird (we could also shift to the center, but the bird is rarely there), mask the image to remove low frequencies, then invert the filtered image and observe it on the magnitude spectrum (take the absolute value). This set of operations is equivalent to a high pass filter, which is useful for edge detection. In the operations considered thus far for Flappy Bird, we have the edges more or less in Figure~\ref{fig:fb_gray}, but by applying the mask we can reduce the problem down to finding the edges of the gap and the signal relative to the position of the bird. As we can see in Figure~\ref{fig:fb_hpf}, this relative signal draws almost a flight path between the bird and the pipes. 

In the Geosigs framework discussed in the paper, we would summarize these images along some interpretation of that flight path. However, with so little structure (two pipes and a bird), there are few examples to consider, but we could consider bird-to-top pipe slope which can be defined as the slope between the bird origin and the most intense point in the path (for example). However, with so little structure and no intention to relate bird-to-top pipe slope at time $t$ to Flappy Bird success as far as we know, we can abstract via convolution and dispense with the additional work done for the transportation problem. 

In the experiments that follow, we will look for evidence that the quality of information has improved. Next, we briefly describe the DQN learning framework. 

\subsubsection{DQN for Flappy Bird}\label{flappy_learner}

{\bf Neural Network Architecture:} Using the pre-processing described in Section~\ref{flappy_inputs}, we train five learners. $Learner_1$ takes as input the sequence of $84\times 84$ gray scale images ($K=4$), uses three CNNs to create representations or features of the data to be used in two fully connected layers as in Table~\ref{table:dqn1}. The last layer is linearly activated to output two values representing jump or not jump. $Learner_2$ and $Learner_3$ differ slightly from $Learner_1$. $Learner_2$ uses $44\times 44$ resized images, changing the input size as noted in Table~\ref{table:dqn2}. $Learner_3$ takes FFT of the $84\times 84$ image as described in the previous section with a $55\times 40$ rectangular mask. $Learner_4$ takes a $65\times 50$ mask. Otherwise, since the size of the image has not changed, Table~\ref{table:dqn1} still applies to learners 3 and 4. $Learner_5$ takes as input the FFT of a $52\times 52$ resized set of images using a $50\times 50$ mask and has a slightly modified neural architecture (fully connected linear layer takes vector of length 576).

\begin{table*}[ht]
	\centering
	\begin{tabular}{cccccccc}
		Layer & Input & Kernel & Stride & Features & Activation & Output \\ 
		\hline
		$CNN_1$ & $84\times 84\times 4$ & 8 & 4 & 32 & ReLU & $20\times 20\times 32$ \\
		\hline
		$CNN_2$ & $20\times 20\times 32$ & 4 & 2 & 64 & ReLU & $9\times 9\times 64$ \\
		\hline
		$CNN_3$ & $9\times 9\times 64$ & 3 & 1 & 64 & ReLU & $7\times 7\times 64$ \\
		\hline
		$Full_1$ & $7\times 7\times 64$ &  &  & 512 & ReLU & 512 \\
		\hline
		$Full_2$ & 3136 &  &  & 2 & Linear & 2 \\
		\hline
	\end{tabular}
	\caption{Layers in neural networks for Learners 1 and 3. } 
	\label{table:dqn1}
\end{table*}

\begin{table*}[ht]
	\centering
	\begin{tabular}{cccccccc}
		Layer & Input & Kernel & Stride & Features & Activation & Output \\ 
		\hline
		$CNN_1$ & $44\times 44\times 4$ & 8 & 4 & 32 & ReLU & $10\times 10\times 32$ \\
		\hline
		$CNN_2$ & $10\times 10\times 32$ & 4 & 2 & 64 & ReLU & $4\times 4\times 64$ \\
		\hline
		$CNN_3$ & $4\times 4\times 64$ & 3 & 1 & 64 & ReLU & $2\times 2\times 64$ \\
		\hline
		$Full_1$ & $2\times 2\times 64$ &  &  & 512 & ReLU & 512 \\
		\hline
		$Full_2$ & 512 &  &  & 2 & Linear & 2 \\
		\hline
	\end{tabular}
	\caption{Layers in neural network for Learner 2. } 
	\label{table:dqn2}
\end{table*}

Since we are primarily interested in comparing learning rates using different inputs, we do not invest in finding the optimal tuning parameters and rely on default optimization settings.

{\bf Deep Q-Learning with Experience Replay:} As the algorithm iterates, it saves transitions to memory that are randomly sampled for future actions (replay memory). A backward pass updates the neural network’s parameters. Actions are executed using $\epsilon$-greedy exploration, meaning that when an action is selected in training, it is either chosen as the action with the highest $Q$-value, or a random action. Choosing between these two actions is random and based on the value of $\epsilon$, and $\epsilon$ is annealed such that lots of random actions are taken initially (exploration), but lots of actions with the maximum $Q$-values will be taken after many iterations (exploitation). The loss function being minimized is
\[
L=\frac{1}{2}\left(\max_{a'}Q(s',a') - Q(s,a) \right)^2.
\]
$Q$-values were calculated using the Bellman equation:
\[
Q(s,a) = \max_a \left( R(s,a) +\gamma Q(s') \right),
\]
where $R(s,a)$ is the reward for action $a$ in state $s$ and $\gamma$ is a discount factor. $\max_{a'}Q(s',a') $ is taken as the maximum output value of the neural network. 

All five learners use the same learning framework, but with varying neural network architecture based on input. In the following section, we compare the different rates at which rewards are accrued. 

\subsubsection{Experimental Results}\label{flappy_results}

In the following experiments, we want to compare training rate and quality of a learner based on different inputs. We do not want neural architecture to vary much nor do we want to train a perfect Flappy Bird algorithm. Instead, we train each learner until the first run that it can accumulate a reward total greater than 10, where each iteration or sequential frame can either gain +0.1 or lose -1.0 if the bird crashes. Note that by choosing 10 with +0.1 per successful frame we are training the bird to get through at least one pipe. 
\begin{table}[ht]
	\centering
	\begin{tabular}{c|c|c}
		Learner & Input & Iterations \\ 
		\hline
		$Learner_1$ & $84\times 84$ & 35,973  \\
		\hline
		$Learner_2$ & $42\times 42$ & NA ($>175,000$) \\
		\hline
		$Learner_3$ & $FFT(84\times 84)$, $55\times 40$ mask & 69,069  \\
		\hline
		$Learner_4$ &  $FFT(84\times 84)$, $65\times 80$ mask & 65,310  \\
		\hline
		$Learner_5$ &  $FFT(52\times 52)$, $50\times 50$ mask & 117,104  \\
		\hline
	\end{tabular}
	\caption{Number of iterations for each learner to successfully earn at least 10 total reward points.} 
	\label{table:fb_learnrates}
\end{table}

Table~\ref{table:fb_learnrates} shows the number of iterations each learner required to hit 10 total reward points for the first time. $Learner_1$ should be seen as the case using the most raw information, so it is not surprising that it trained in the smallest number of iterations. Resizing the input down to $42\times 42$ for $Learner_2$ shows how much information loss impacts learning, since 10 reward points could not be earned by 175,000 iterations. Using compression with a small mask pivoted on the bird, $Learner_3$ reached 10 points at about 69,000 iterations, almost twice that of $Learner_1$. By increasing the rectangular mask as input for $Learner_4$, that number of iterations was reduced down to 65,310. $Learner_5$ uses a smaller filtered image, but earns the 10 points by about 110,000 iterations.

\begin{figure}[h]
	\centering
	\includegraphics[height=3in,width=3in]{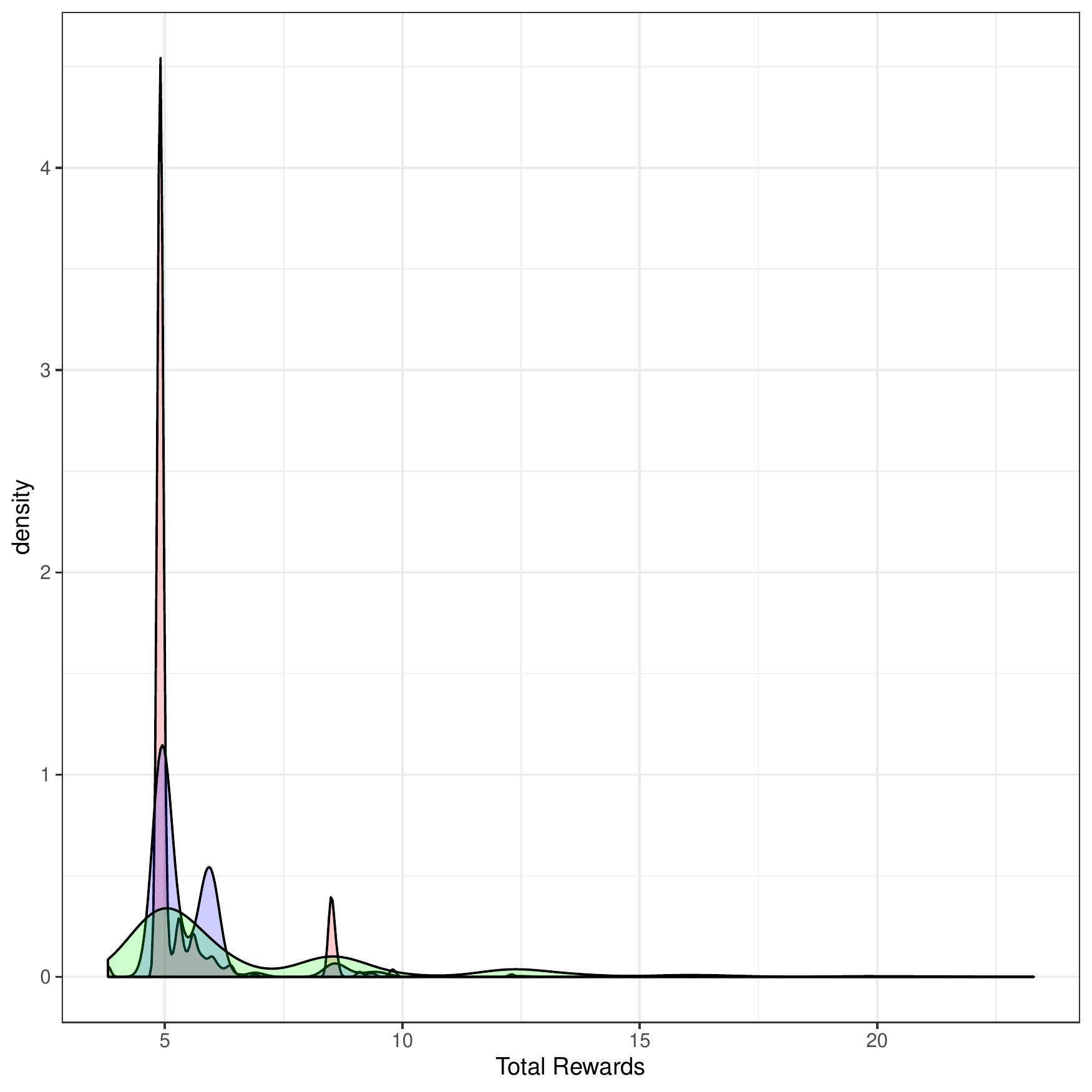}
	\caption{Distribution of rewards for each of the learners ($Learner_3$: red, $Learner_4$: blue, $Learner_5$: green), except $Learner_1$, which was just concentrated at 4.9 (mean=median=max=4.9). }
	\label{fig:fb_rewards}
\end{figure}

Next, using the fitted learners that hit 10 total points, we simulate 500 runs of Flappy Bird for each to play. $Learner_1$ completed the 500 runs with total rewards concentrated at 4.9 (mean=median=max=4.9), totaling up to 2,449.9 points across the 500 games. This suggests that this learner did not have a diverse set of moves. As seen in Figure~\ref{fig:fb_rewards}, $Learner_3$ also tended to lean towards 4.9 total rewards, but earned 2,681 total points. Earning 2,753.6 points, we observe that $Learner_4$ was bi-modal: concentrated at 4.9 and 5.9. $Learner_5$ proved to be the most successful: a better distribution across rewards and earned 3,412.7 points. This shows that as we move from $Learner_1$ to $Learner_3$ and on, the algorithm learned how to better play the game. This improvement is subject to the number of iterations, the quality of the input information, and also some uncertainty based on realizations of the games. We expect algorithms with more iterations to perform better, but $Learner_4$ outperformed $Learner_3$ and this may be due to the quality of the information revealed by the FFT. Of course, we do not achieve dimension reduction by using $Learner_3$ or $Learner_4$, but $Learner_5$ outperformed all models even though it had the smallest input to work with. Again, this demonstrates the value of filtering in the pre-preprocessing step. 

As for uncertainty, 500 game plays may not be enough to rule that $Learner_4$ is better than $Learner_3$. However, this paper is about what can be achieved using image transformations in a learning environment. By filtering the Flappy Bird image with a mask pivoted on the bird center of the screen, we can reduce the size of the input and while it takes longer to achieve the same total reward as another algorithm, it learns quality information on how to play the game. This tailoring via image compression is what drives us to consider understanding what is important about an image. In the case of Flappy Bird, it is straightforward: there is a game display and a constant structure of a bird and pipes somewhere on the screen. For the transportation problem in the paper, it was geographic information that could be context-specific for the region. Nevertheless, as RL problems, these have in common the value-add of quality information. 

\bibliography{myarticles}
\bibliographystyle{plain}

\end{document}